%% file: main.tex
\pdfoutput=1

\documentclass[11pt]{article}

\usepackage[]{acl}

\usepackage{times}
\usepackage{latexsym}

\usepackage[T1]{fontenc}

\usepackage[utf8]{inputenc}

\usepackage{color}
\usepackage{tabularx}
\usepackage{tcolorbox}
\usepackage{xcolor}
\usepackage{colortbl}
\usepackage{booktabs}
\usepackage{multirow}

\usepackage{graphicx} 
\usepackage{float}
\usepackage{subfigure} 
\usepackage{amssymb}

\usepackage{hyperref}
\usepackage{tablefootnote}
\usepackage{xspace}

\usepackage{float}
\usepackage{amsmath}
\usepackage{bm}
\usepackage{algorithmicx}
\usepackage{algorithm}
\usepackage{algpseudocode}

\usepackage{arydshln}

\usepackage{amssymb}
\usepackage{pifont}
\usepackage{enumitem}
\usepackage{fontawesome5}

\usepackage{microtype}

\newcommand{\ours}{\textsc{DocMath-Eval}\xspace}

\newcommand{\dev}{testmini\xspace}

\newcommand{\spe}{specialized\xspace}
\newcommand{\nmodel}{48\xspace}
\newcommand{\nexample}{4,000\xspace}
\newcommand{\ourss}{\textsc{DM}$_{\text{SimpShort}}$\xspace}
\newcommand{\oursl}{\textsc{DM}$_{\text{SimpLong}}$\xspace}
\newcommand{\ourcs}{\textsc{DM}$_{\text{CompShort}}$\xspace}
\newcommand{\ourcl}{\textsc{DM}$_{\text{CompLong}}$\xspace}

\newcommand{\norg}{17\xspace}

\newcommand{\eg}{\hbox{\emph{e.g.,}}\xspace}
\newcommand{\ie}{\hbox{\emph{i.e.,}}\xspace}

\title{\ours: Evaluating Math Reasoning Capabilities of LLMs \\in Understanding Long and Specialized Documents}

\author{Yilun Zhao\thanks{~~Equal Contributions.}~~$^{1}$ \quad Yitao Long$^{*~2}$ \quad Hongjun Liu$^{2}$ \quad Ryo Kamoi$^{3}$ \quad Linyong Nan$^{1}$\\ \bf{Lyuhao Chen$^{4}$ \quad Yixin Liu$^{1}$ \quad Xiangru Tang$^{1}$ \quad Rui Zhang$^3$ \quad Arman Cohan$^{1,5}$} \vspace{4pt}\\
$^1$Yale University \quad $^2$New York University \quad $^3$Penn State University \\ $^4$Carnegie Mellon University \quad $^5$Allen Institute for AI\\\newline \vspace{10pt}
}

\begin{document}
\maketitle

\begin{minipage}[t]{2\linewidth}
\vspace{-1.75cm}
  \centering
  \href{https://github.com/yale-nlp/DocMath-Eval}{{\faGithub{}}\xspace\texttt{https://github.com/yale-nlp/DocMath-Eval}} \\\vspace{2pt}
  \href{https://docmath-eval.github.io}{{\faGlobe{}}\xspace\texttt{https://docmath-eval.github.io/}} \\
\vspace{0.5cm}
\end{minipage}

\begin{abstract}
\input{main/0-abstract}
\end{abstract}

\input{main/1-introduction}
\input{main/2-related_work}
\input{main/3-task}
\input{main/4-experiment}
\input{main/6-conclusion}
\input{main/limitations}
\input{main/acknowledgement}

\bibliography{custom, anthology, llms}

\appendix

\input{appendix/main}

\end{document}

%% file: main/0-abstract.tex
Recent LLMs have demonstrated remarkable performance in solving exam-like math word problems. However, the degree to which these numerical reasoning skills are effective in real-world scenarios, particularly in expert domains, is still largely unexplored.
This paper introduces \ours, a comprehensive benchmark specifically designed to evaluate the numerical reasoning capabilities of LLMs in the context of understanding and analyzing \spe documents containing both text and tables. 
We conduct an extensive evaluation of \nmodel LLMs using Chain-of-Thought and Program-of-Thought prompting techniques, aiming to comprehensively assess the capabilities and limitations of existing LLMs in \ours. 
We found that even the current best-performing system (\ie GPT-4o)
still significantly lags behind human experts in solving complex numerical reasoning problems grounded in long contexts.
We believe that \ours can serve as a valuable benchmark for evaluating LLMs' capabilities in solving challenging numerical reasoning problems within expert domains.

%% file: main/1-introduction.tex
\section{Introduction}
\input{figures/main_example_txt}

Recent advancements in large language models (LLMs) have attracted significant attention due to their capabilities in solving a broad range of tasks~\cite{OpenAI2023GPT4TR, llama3modelcard}, including math word problems (MWPs) commonly found in academic exams~\cite{wang-etal-2017-deep,miao-etal-2020-diverse,amini-etal-2019-mathqa, cobbe2021training, hendrycks2021measuring, cobbe2021training, lu2023dynamic, chen-etal-2023-theoremqa}. These MWPs vary from basic arithmetic to advanced algebra, showcasing LLMs' proficiency in numerical reasoning — a crucial skill for interpreting and manipulating numerical data across various contexts. 
Despite this progress, there is still a significant gap in understanding the practicality of LLMs' numerical reasoning in real-world scenarios, particularly in specialized fields such as finance, medicine, and science. As illustrated in \autoref{fig:example}, these expert domains necessitate LLMs to interpret complex, domain-specific documents, applying numerical reasoning to complex problem-solving~\cite{chen-etal-2021-finqa, zhu-etal-2021-tat, zhao-etal-2022-multihiertt, li-etal-2022-learning}.
Recognizing this gap, our research focuses on the finance domain~\cite{li-etal-2022-finmath, wu2023bloomberggpt, yang2023fingpt, callanan2023gpt, xie2024finben}. The finance industry often deals with lengthy and data-intensive documents that demand advanced numerical reasoning skills for accurate analysis and decision-making.

We introduce \textbf{\ours}, 
a comprehensive and standardized benchmark that systematically evaluates the numerical reasoning capabilities of LLMs in understanding and interpreting \spe documents containing both textual and tabular data. \ours encompasses four evaluation sets, each with varying levels of difficulty in \emph{numerical reasoning} and \emph{document understanding}. 
Specifically, We construct a new evaluation set, \textbf{\ourcl}, from scratch, to examine the LLM's capabilities in performing \emph{complex} numerical reasoning over \emph{extreme long} documents containing \emph{multiple} tables. We also adapt and re-annotate four existing finance QA benchmarks to develop three additional, less challenging evaluation sets: 1) \textbf{\ourss} based on TAT-QA~\cite{zhu-etal-2021-tat} and FinQA~\cite{chen-etal-2021-finqa}, necessitates \emph{simple} numerical reasoning over \emph{short} document with \emph{one} table; 
2) \textbf{\oursl} based on MultiHiertt~\cite{zhao-etal-2022-multihiertt}, necessitates \emph{simple} numerical reasoning over \emph{long} document with \emph{multiple} tables; 
and 3) \textbf{\ourcs} based on TAT-HQA~\cite{li-etal-2022-learning}, necessitates \emph{complex} numerical reasoning over \emph{short} document with \emph{one} table. 

We conduct an extensive evaluation on \ours, covering a total of \nmodel proprietary and open-source LLMs from \norg organizations. 
Two prompting methods, Chain-of-Thought (CoT)~\cite{wei2022chain} and Program-of-Thought (PoT)~\cite{chen2023program}, are applied for result analysis. 
Our experimental results indicate that while the existing best-performing LLM on average (\ie GPT-4o) can achieve high performance in simple settings (\eg \ourss), it still falls short of human experts in more challenging ones, \ie, \ourcl. 
Moreover, Claude-3.5-Sonnet outperforms other LLMs, achieving an accuracy of 40.0\% on the \ourcl set when applying CoT prompting. 
However, it still lags far behind human expert performance, which stands at 76\%.
This significant gap between LLMs and human experts underscores the challenges presented by \ours. It underscores the importance of advancing LLMs' numerical reasoning and document understanding abilities to effectively apply them in the real-world specialized domains.

We conclude our main contributions as follows:
\begin{itemize} [leftmargin=*]
\itemsep0em 
\item We introduce \ours, a comprehensive benchmark designed to systematically evaluate LLMs' numerical reasoning ability to understand and interpret long and \spe documents. This includes a newly developed, challenging evaluation set and three adapted evaluation sets for varying difficulty levels.
\item We conduct an extensive evaluation encompassing a wide range of LLMs, including those specialized in math and coding. We also incorporate different prompting methods (\ie CoT and PoT) to comprehensively assess the capabilities and limitations of existing LLMs in our task.
\item Our experimental results reveal a noticeable performance gap compared to human experts in more complex scenarios (\ie problems requiring complex numerical reasoning over long documents). This highlights the limitations of current LLMs in complex real-world applications and the need for continued advancements.
\end{itemize}

\input{tables/data_statistics}

%% file: figures/main_example_txt.tex
\begin{figure}[!t]
    \centering
    \includegraphics[width = \linewidth]{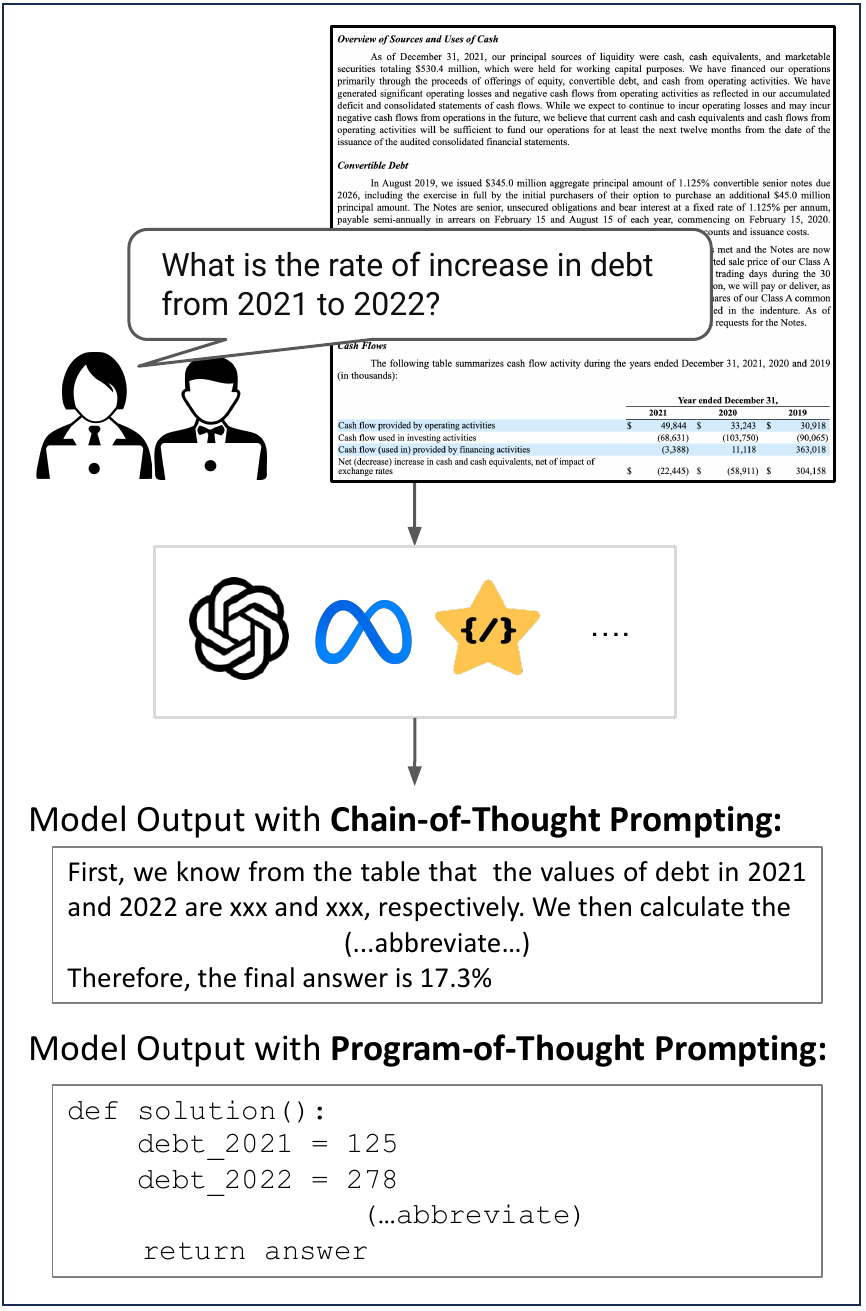}
    \caption{The overview of \ours and the prompting methods explored. \ours evaluates the LLMs' performance in the context of understanding and analyzing financial documents containing both text and tables. The models are required to first locate question-relevant data points within lengthy documents, and then apply numerical reasoning and specialized financial knowledge to answer the question.}
    \label{fig:example}
\end{figure}

%% file: tables/data_statistics.tex
\begin{table*}[t!]
\centering 
\addtolength{\tabcolsep}{-0.2em}
\resizebox{\linewidth}{!}{
\begin{tabular}{lrrrr}
\toprule
\textbf{~~~Property} \texttt{(Median/Avg)}  & \textbf{\ourss} & \textbf{\oursl} & \textbf{\ourcs} &  \textbf{\ourcl (new)}\\
\midrule
~~~\multirow{2}{*}{Data Source} & TAT-QA~\cite{zhu-etal-2021-tat} & MultiHiertt & TAT-HQA & expert annotated \\
& FinQA~\cite{chen-etal-2021-finqa} & \cite{zhao-etal-2022-multihiertt} & \cite{li-etal-2022-learning} & from scratch\\
\midrule
~~~Question Length   & 19 / 20.0  & 21 / 21.6 & 29 / 30.1 & 34 / 37.7 \\

\midrule

\# Sentences in Text  & 14 / 16.9 & 64 / 66.9 & 6 / 7.8 &  \textcolor[RGB]{0, 107, 35}{535 / 752.3}\\
\# Words in Text  & 504 / 506.6 & 2,216 / 2,334.0 & 251 / 314.2 & \textcolor[RGB]{0, 107, 35}{25,149 / 34,589.0} \\

\midrule

\# Table   &  1 / 1.0 & 4 / 3.9 & 1 / 1.0 & \textcolor[RGB]{0, 107, 35}{46 / 72.5}\\
\# Rows per Table  & 6 / 7.0 & 9 / 11.6 & 7 / 8.2 & 3 / 7.5\\
\# Columns per Table  &  5 / 4.7 & 4 / 4.5 & 5 / 5.0 & 3 / 3.1\\

\midrule

\# Text Evidence  & 0 / 0.4 & 1 / 0.9 & 0 / 0.4 & 1 / 1.0\\
\# Table Evidence  & 1 / 0.9 & 1 / 1.1 & 1 / 1.0 & 1 / 1.0\\
\% Questions $w.$ Table Evidence & 92.9\% & 86.4\% & 97.8\% & 76.3\%\\
\midrule
\# Math Operations in Python Solution  & 2 / 2.1 & 2 / 2.3 & 2 / 2.3 & \textcolor[RGB]{35, 71, 178}{4 / 4.9}\\
\# Code Lines in Python Solution  &  5 / 5.3 & 6 / 5.9 & 5 / 5.3 & \textcolor[RGB]{35, 71, 178}{8 / 8.2}\\
\# Comment Lines in Python Solution   &  2 / 2.0 & 2 / 2.0 & 2 / 2.0 & \textcolor[RGB]{35, 71, 178}{2 / 3.4}\\
\midrule
Development set & 200 & 100 & 200 & 300\\
Test set & 800 & 400 & 800 & 1,200\\
Total Size & 1,000 & 500 & 1,000 & 1,500 \\
\toprule
\end{tabular}
}
\caption{Basic statistics of \ours dataset. Our newly constructed evaluation set, \ourcl, poses unique challenges in both \textcolor[RGB]{35, 71, 178}{numerical reasoning} and \textcolor[RGB]{0, 107, 35}{financial document understanding}.}
\label{tab:data_statistics}
\end{table*}

%% file: main/2-related_work.tex
\section{Related Work}
\paragraph{Math Word Problems}
The research community has shown significant interest in the vital role of numerical reasoning skills in LLMs. These skills are vital for models to effectively engage in complex problem-solving. To this end, a wide variety of MWP datasets have been proposed in recent years~\cite{hosseini-etal-2014-learning,  koncel-kedziorski-etal-2016-mawps, wang-etal-2017-deep, ling-etal-2017-program, cobbe2021training}. More challenging datasets have recently been introduced to enhance diversity~\cite{miao-etal-2020-diverse}, difficulty~\cite{chen-etal-2023-theoremqa, hendrycks2021measuring}, and adversarial robustness~\cite{patel-etal-2021-nlp}. However, existing MWP datasets predominantly focus on problems akin to academic exams, with a limited emphasis on real-world scenarios. 
Addressing this gap, our paper introduces a novel and comprehensive benchmark designed to evaluate LLMs' abilities in understanding and interpreting long and specialized documents through numerical reasoning.

\paragraph{Numerical Reasoning over Documents}
Numerical reasoning over documents requires models to have a deep understanding of context and the ability to derive answers through numerical reasoning~\cite{dua-etal-2019-drop}. Applying these models in the finance domain~\cite{Xie2023PIXIUAL, wu2023bloomberggpt, yang2023fingpt} presents additional challenges in terms of interpreting hybrid data~\cite{zhu-etal-2021-tat} and utilizing domain-specific expertise~\cite{chen-etal-2021-finqa, zhao2023knowledgemath}. 
Numerous datasets focusing on numerical reasoning over specialized documents have been proposed recently. Two notable benchmarks are TAT-QA~\cite{zhu-etal-2021-tat} and FinQA~\cite{chen-etal-2021-finqa}, which represent pioneering efforts in studying numerical reasoning in finance, particularly requiring the fusion of tabular and textual content.  Building upon TAT-QA, a more challenging dataset named TAT-HQA~\cite{li-etal-2022-learning} was developed, focusing on counterfactual questions in relation to the provided context. Additionally, MultiHiertt~\cite{zhao-etal-2022-multihiertt} focuses on numerical reasoning over longer financial documents containing multiple tables.
However, as illustrated in \autoref{tab:data_statistics}, these four datasets focus on less challenging scenarios, where either simple numerical reasoning (\eg calculating the increasing rate or average value) is sufficient, or the input context is short. Furthermore, there is a lack of a standardized benchmark for systematically evaluating models' performance across varying difficulty levels in terms of numerical reasoning and document understanding.

%% file: main/3-task.tex
\section{\ours}
In this section, we first offer a formal definition of the \ours task. We then explain the rationale and methodology for adopting Python program as the standardized solution format for \ours. Subsequently, we detail the data annotation process used to construct the challenging \ourcl evaluation set, as well as the data re-annotation process for compiling the other three evaluation sets. 
\autoref{tab:annotator_profiles} in the Appendix presents the profiles of the seven annotators involved. 
Finally, we present human-level performance on each evaluation set in \ours. 

\subsection{Task Formulation}
We formally define the task of \ours in the context of LLMs as follows: Presented with a numerical reasoning question $q$ and a financial document consisting of textual contents $E$ and structured tables $T$, the task is to generate the numeric-value answer $a$:
\begin{equation}
\label{eq:formulation}
    \hat{a} = \arg\max_a P_{\mathbf{LM}}(a~|~q, E, T)
\end{equation}
To obtain the best candidate answer $\hat{a}$, we use greedy decoding in all our LLM evaluations. 

\subsection{Solution Format Standardization}\label{sec:program}
We observe that existing finance QA datasets feature solutions in various formats. Specifically, TAT-QA~\cite{zhu-etal-2021-tat} and TAT-HQA~\cite{li-etal-2022-learning} utilize text, while MultiHiertt~\cite{zhao-etal-2022-multihiertt} employs mathematical expressions, such as \texttt{100/3}, and FinQA~\cite{chen-etal-2021-finqa} uses math programs, such as \texttt{divide(100,3)}, for solution annotations. This diversity in annotation formats hinders the development of a unified evaluation framework to assess LLM performance across different benchmarks. 
Additionally, text-based solutions often fall short in precision and clarity, making them less suitable for computational problem-solving; and the solutions presented as mathematical equations or programs can be less descriptive, with the intended semantic meaning of the equations sometimes being unclear.

To overcome the aforementioned limitations, in \ours, we represent solutions using Python programs~\cite{zhao2023knowledgemath}.
Such a unified Python program format supports a standardized and effective evaluation framework for LLM assessment.
Specifically, annotators are instructed to initially define variables at the start of the Python function, beginning with ``\texttt{def solution():}''. These variables should align with the primary elements or quantities referenced in the question or relevant content in the documents.
They then write a Python program that methodically address the problem, solving it step by step. Additionally, annotators receive a bonus for writing detailed comments, thereby enhancing the code's readability and understandability.
To verify the correctness and performance of the solutions, our annotation interface automatically runs the Python function. This process checks that the output is either a float or int and ensures that the execution finishes without any errors.

\subsection{Data Re-Annotation From Public Datasets}
\label{sec:reannotation}
We re-annotate four existing datasets and incorporate them into \ours. Specifically, we re-annotate TAT-QA~\cite{zhu-etal-2021-tat} and FinQA~\cite{chen-etal-2021-finqa} for \ourss, MultiHiertt~\cite{zhao-etal-2022-multihiertt} for \oursl, and TAT-HQA~\cite{li-etal-2022-learning} for \ourcs.

\paragraph{Question Validation and Re-annotation} 
We instruct the annotators to identify and remove questions with incorrect annotations or those whose answers are not numerical. Annotators are then asked to enhance each question by adding a scale descriptor to ensure clarity and specificity. For example, \textit{"Question: What is the average payment volume per transaction for American Express? \textbf{(in billions)}"}. They were also asked to correct any identified errors in the original questions.

\paragraph{Solution Validation and Re-annotation} 
As outlined in Section~\ref{sec:program}, we require annotators to rewrite the original solutions into a unified Python format, standardizing variable names and adding comments to enhance the readability of the solutions. 
Regarding the supporting evidence annotation, we initially convert the original evidence annotations to our format. We then highlight these evidences in the annotation interface, and direct annotators to verify their correctness.

\subsection{Data Annotation From Scratch}
In real-world scenarios, financial professionals typically need to handle documents spanning tens of pages, along with problems that require more complex numerical reasoning combined with financial knowledge. 
However, as previously discussed, existing benchmarks~\cite{zhu-etal-2021-tat, chen-etal-2021-finqa, zhao-etal-2022-multihiertt, li-etal-2022-learning} focus on less challenging scenarios, where either simple numerical reasoning is sufficient, or the input context is short. To bridge this gap, we have developed a new, challenging evaluation set,  \textbf{\ourcl}, from scratch. This set focuses on settings that more closely align with real-world scenarios, where models are required to perform complex numerical reasoning over long financial documents for problem solving. The annotation process is as follows:

\paragraph{Source Document Collection} 
Following previous work~\cite{zhu-etal-2021-tat, chen-etal-2021-finqa, zhao-etal-2022-multihiertt}, we use the quarterly (i.e., Form 10-Q) and annual reports (i.e., Form 10-K) of companies as our source documents, which are publicly available at 
the open-source database\footnote{\url{https://www.sec.gov/edgar/search/}} of U.S. Securities and Exchange Commission.
After collecting all the source documents, we utilize a commercial API\footnote{\url{https://sec-api.io/}} to extract their textual and tabular content. Subsequently, we apply a heuristic-based method to preprocess these two formats of content. The preprocessed documents are then passed to expert annotators for question annotation.

\paragraph{Data Annotation} \label{sec:annotate_new}
Given a financial document, annotators are first required to briefly read its content and determine the data points to be used in the question. They must then compose the question and highlight the selected paragraphs or tables as evidence supporting it. 
Finally, the annotators are required to write down the solution to the question in Python program format, as discussed in Section~\ref{sec:program}.
We set up a \emph{bonus payment system} for complex annotations that involve difficult document comprehension and numerical reasoning. Specifically, to increase the difficulty of document understanding, we award bonuses to annotators for questions that necessitate information from: 1) multiple tables, 2) multiple sections, or 3) a combination of tables and textual content. To enhance the challenge in numerical reasoning, we provide bonuses for questions requiring financial expertise or involving complex mathematical operations. 
If such annotations are validated during the quality validation stage, a bonus payment will be added.

\paragraph{Quality Validation}
We implement a comprehensive quality validation protocol to ensure that each annotated example meets the required standards. For every question annotation, we assign it to another annotator, recognized for their high performance in annotation, to verify its accuracy. 
This process involves manually locating the question-relevant evidence in the documents using our retrieval-based search toolkits. They then compare this evidence with the original annotations and correct any errors found. Additionally, validators are tasked with confirming the accuracy of the annotated solutions. 
We offer bonus payments to annotators for identifying erroneous annotations. Ultimately, 232 of the annotated questions are flagged as erroneous and are subsequently revised.
\autoref{tab:annotation_aggrement} in the Appendix presents the human evaluation scores and inter-evaluator agreements for a subset of 200 sampled examples. \ours exhibits superior annotation quality and a high degree of inter-annotator agreement. 

\subsection{Expert-level Performance Evaluation}
To give a general yet insightful estimate of the performance on each of the \ours sets, we enlisted two professionals who hold Chartered Financial Analyst licenses to conduct the evaluation.
Regarding human expert performance on \ourss and \oursl, we report the same results as those in the original papers, with accuracy of 91\% and 87\%, respectively.
For \ourcs and \ourcl, We randomly sample 25 examples from each set, asking the expert evaluators to answer the questions individually within a four-hour period. They achieve accuracy of 88\% and 80\% on \ourcs (average 84\%); and accuracy of 72\% and 80\% on \ourcl (average 76\%).

\subsection{Dataset Release}
\autoref{tab:data_statistics} presents the data statistics of four developed evaluation sets. \ours contains a total of \nexample questions with high-quality annotations, featuring varying difficulty levels in numerical reasoning and document understanding.
We randomly partitioned the dataset into two subsets: \emph{\dev} and \emph{test}. The \emph{\dev} subset includes 800 examples and is intended for model development and validation. The \emph{test} subset consists of the remaining 3,200 examples, which are reserved for standard evaluation. To avoid data contamination~\cite{deng2024survey}, the features directly related to the ground truth for the \emph{test} set are kept private. Instead, we have developed and manage an online evaluation platform, where researchers can assess models and participate in a leaderboard.

%% file: main/4-experiment.tex
\input{figures/CoT_prompt_example}
\section{Experiment Setup}
This section discusses the experiment setup, including the evaluated LLMs, prompting methods, and our implementation details.

\subsection{Evaluated Large Language Models}
Our goal is to investigate the capabilities of current state-of-the-art LLMs on \ours to better understand their strengths and limitations. To this end, we evaluate a wide range of models, including 32 general-purpose LLMs, 4 math-specific LLMs, 6 code-based LLMs, and 7 mixture of experts (MoE) models. 
The specific details of each evaluated LLM, including the exact version used, can be found in \autoref{tab:model-detail} in the Appendix.







\subsection{Prompting Methods}
Following recent works on LLM reasoning benchmarks~\cite{lu2024mathvista, chen-etal-2023-theoremqa}, we evaluate two commonly used prompting methods for math reasoning:

\paragraph{Chain-of-Thought} The CoT method~\cite{wei2022chain} instructs the LLMs to explicitly outline their reasoning process step by step before arriving at the final answer. \autoref{fig:cot_example} presents the CoT prompt used in our experiment.

\paragraph{Program-of-Thought} The PoT method~\cite{chen2023program} separates computation from the reasoning process by instructing the LLMs to produce a structured program that encapsulates the reasoning steps. The final answer is obtained by executing the generated program. \autoref{fig:pot_example} in Appendix presents the PoT prompt we used.

\input{tables/testmini_results}
\subsection{Implementation Details}
\paragraph{LLM Experiment}
The experiments involving open-sourced LLMs were conducted using the \texttt{vLLM} framework~\cite{kwon2023efficient}. In all the experiments, we used a temperature setting of 1.0 and maximum output length of 512. 
Given the extensive context length of input document, the main evaluation of \ours is conducted under a \emph{zero-shot} setting, aiming to assess LLMs' capabilities to generate accurate answers without few-shot demonstrations or additional training. 

\paragraph{Input Tabular Data Serialization}
Building on previous work that evaluated LLMs on table-relevant tasks~\cite{chen-2023-large, zhao-etal-2023-qtsumm, zhao-etal-2023-investigating}, we present our method for processing tabular data in documents. Specifically, we separate headers or cells in different columns using a vertical bar (|), and rows using a newline. This approach allows for the direct feeding of flattened table input into LLMs. 
In our preliminary study, we found that most LLMs can comprehend these table formats well. Nevertheless, we believe that future research could explore more effective methods for encoding tabular data~\cite{fang2024large}.

\paragraph{RAG-based Setting for \ourcl} \label{sec:retriever}
For the \ourcl subset, the input document length is extremely long and exceeds the context length limit of evaluated LLMs. Therefore, in our main experiments with \ourcl, we evaluate models using the retrieval-augmented generation (RAG) setting. 
In this setting, external retrievers are employed to extract the top-$n$ most relevant textual and tabular evidence from the source document. 
We maintain the original relative order of the evidence and input it into the LLMs to answer the given question.
We experiment with commonly-used sparse retriever, \ie BM25~\cite{robertson1995okapi}, 
and three dense retrievers, including OpenAI Embedding 3 small \& large versions~\cite{neelakantan2022text} and Contriever~\cite{izacard2022unsupervised}.

\paragraph{Final Answer Extraction}
For LLMs using CoT prompting, we adopt the answer extraction process developed by \citet{chen-etal-2023-theoremqa} to extract the final answer from the model's output. For LLMs employing PoT prompting, we first develop a heuristic method to extract the generated python solution from the model response. We then execute it to obtain the final answer. 

\section{Results and Analysis}
We next discuss our main findings from the experiments and our analysis of the \ourcl subset.
\subsection{Main Results}
\autoref{tab:testmini} and \autoref{tab:test} in the Appendix present the LLM performance on the \ours testmini and test sets, respectively.

While the current best-performing LLM, GPT-4o, achieves performance comparable to human experts in simple problem settings (\ie \ourss and \ourcs), we find significant performance gaps in more challenging settings. Specifically, GPT-4o achieves an accuracy of 41.0\% on \ourcl with PoT, which is far behind the human expert performance of 76.0\%. This underscores the need for ongoing LLM development, particularly in complex problem-solving over long and specialized documents.
Most open-source LLMs still lag behind the proprietary LLMs. 
However, the two DeepSeek-V2-* models come close to matching the performance of the leading proprietary models. The DeepSeek-V2 even outperforms GPT-4o on the \ourcl subset.
This suggests that open-source LLMs have the potential to bridge the performance gap with the leading proprietary models in the near future.

The code-specific and proprietary LLMs generally perform as well as or better with PoT prompting compared to CoT prompting. This is likely because LLMs are prone to making errors during complex mathematical computations, as revealed in concurrent work~\cite{zhao2023knowledgemath}.
Additionally, for math-specific LLMs, InternLM2-Math-Plus outperforms its base model in CoT performance, with average accuracy rising from 9.9\% to 13.0\%. This highlights the impact of instruction-tuning in improving math reasoning abilities.

\subsection{Analysis on \ourcl Set}
We next conduct a detailed analysis of the RAG setting, long-context LLMs, and model failure cases.

\input{tables/retrieval}
\paragraph{RAG Analysis}
We analyze the impact of retriever performance on the final accuracy of RAG-based LLM systems by selecting the Llama-3-70B and GPT-4o models for our study.
As demonstrated in \autoref{tab:retriever}, the OpenAI Embedding-3 significantly outperforms Contriever and BM25. 
Additionally, improved retriever performance consistently boosts the final accuracy of the models in our task. These results highlight the need for future work to develop more advanced information retrieval techniques for enhancing complex problem-solving over long and specialized documents.

\input{tables/long_context}
\paragraph{Long-Context LLM Analysis}
In addition to using RAG for analyzing long specialized documents, recent advancements have extended the input length of LLMs to handle lengthy documents~\cite{su2023roformer}. 
We compare models with a context length limit of over 100K under both the RAG (as used in the main results) and Long-Context settings, where the entire document is input.
As illustrated in \autoref{tab:long-context-results}, the evaluated models generally achieve close performance under RAG and long-context settings. 
This indicates that models with extended context lengths can effectively process lengthy inputs without a significant drop in performance compared to the RAG setting.

\input{tables/case_study}
\paragraph{Error Analysis}
To better understand the strengths and weaknesses of LLMs, we conduct an extensive error analysis. 
This analysis focuses on 100 randomly selected examples from the \ourcl testmini set where GPT-3.5-turbo failed. 
We identify four common types of errors in current LLMs: inaccurate evidence retrieval, calculation errors, table misunderstandings, and exceeding context length. 
A detailed explanation for each type is provided in \autoref{tab:casestudy}. 

%% file: figures/CoT_prompt_example.tex
\begin{figure}[!t]
\includegraphics[width = \linewidth]{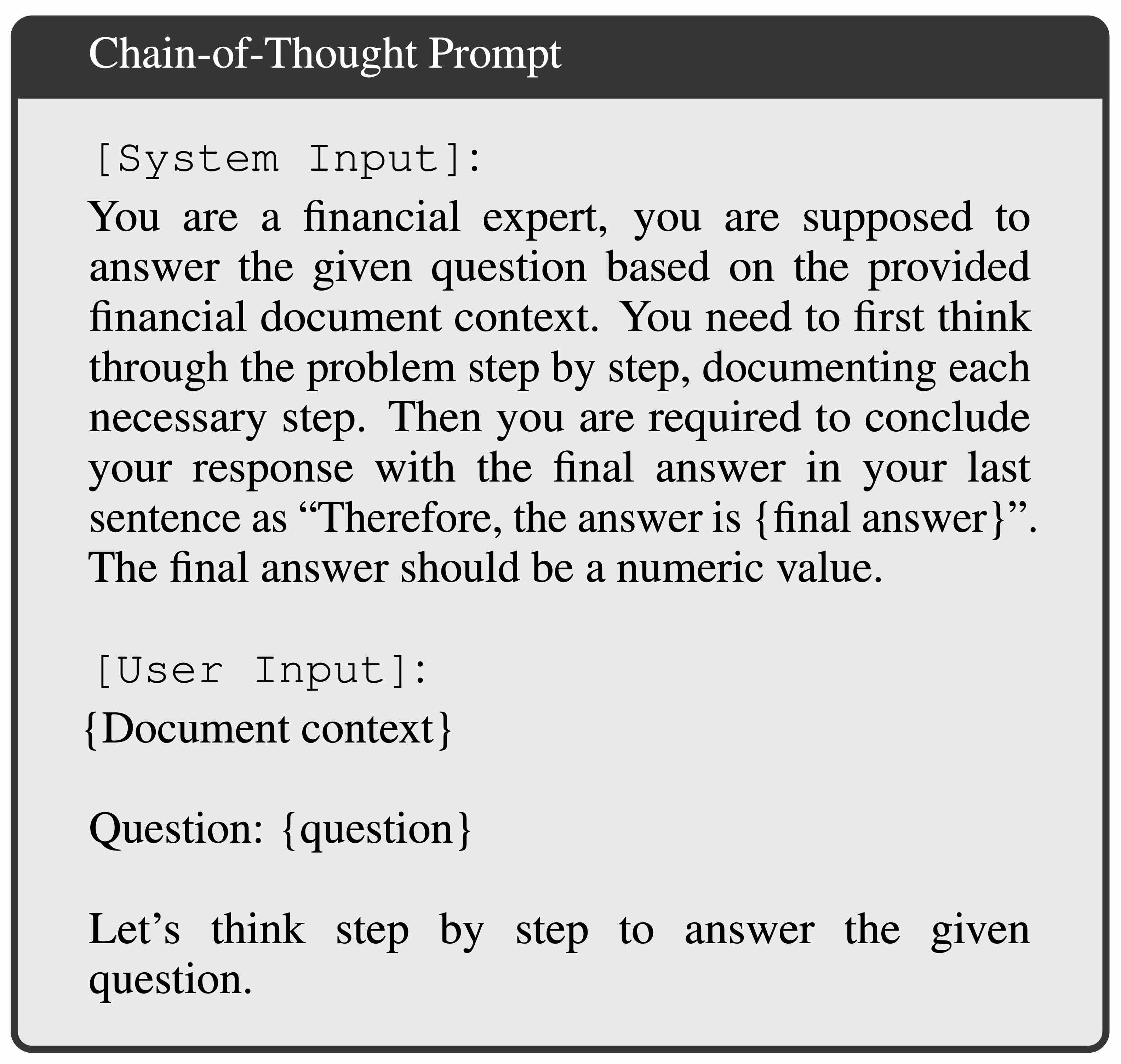}

\caption{Example of \emph{zero}-shot CoT prompt used.}
\label{fig:cot_example}
\end{figure}

%% file: tables/testmini_results.tex
\begin{table*}[!t]
\centering
\resizebox{0.95\textwidth}{!}{%
\addtolength{\tabcolsep}{0.1em}
\begin{tabular}{lrlrrrrrrrrrrrrrrr}
\toprule
\multirow{2}{*}{\textbf{Model}} & \multirow{2}{*}{\textbf{Size}} & \multirow{2}{*}{\textbf{Notes}} & \multicolumn{2}{c}{\textbf{\ourss}} && \multicolumn{2}{c}{\textbf{\ourcs}} && \multicolumn{2}{c}{\textbf{\oursl}} && \multicolumn{2}{c}{\textbf{\ourcl}} && \multicolumn{2}{c}{\textbf{Avg. Acc}} \\
\cmidrule(lr){4-5} \cmidrule(lr){7-8} \cmidrule(lr){10-11} \cmidrule(lr){13-14} \cmidrule(lr){16-17}
& & & PoT & CoT && PoT & CoT && PoT & CoT && PoT & CoT &&  PoT & CoT\\
\midrule

Human Expert & & & \multicolumn{3}{c}{91.0} & \multicolumn{3}{c}{87.0} & \multicolumn{3}{c}{84.0} & \multicolumn{3}{c}{76.0} &\\
\midrule

\multicolumn{18}{c}{\emph{\textbf{Proprietary LLMs}}} \\\noalign{\vskip 0.5ex}

GPT-4o &  &  &84.0 & \cellcolor{red!35}86.0 &  & 69.5 & \cellcolor{red!5}76.5 &  & \cellcolor{red!20}56.0 & \cellcolor{red!35}64.0 &  & \cellcolor{red!5}\underline{41.0} & 36.7 &  & 60.8 & \cellcolor{red!35}62.4 \\
GPT-4-Turbo &  &  &\cellcolor{red!5}\underline{85.5} & \cellcolor{red!20}82.5 &  & \cellcolor{red!5}80.0 & \cellcolor{red!35}81.0 &  & \cellcolor{red!5}\underline{56.0} & 53.0 &  & \underline{38.7} & \cellcolor{red!5}38.3 &  & \cellcolor{red!20}\underline{62.9} & \cellcolor{red!20}61.9 \\
Claude-3-Opus &  &  &\underline{80.5} & 79.5 &  & 73.5 & \cellcolor{red!20}77.5 &  & 51.0 & \cellcolor{red!5}61.0 &  & \cellcolor{red!20}\underline{42.0} & \cellcolor{red!20}39.7 &  & 60.6 & \cellcolor{red!5}61.8 \\
Claude-3.5-Sonnet &  &  &\underline{78.0} & 77.0 &  & \underline{76.0} & 69.5 &  & 54.0 & \cellcolor{red!20}61.0 &  & \cellcolor{red!35}\underline{44.0} & \cellcolor{red!35}40.0 &  & \underline{61.8} & 59.2 \\
Claude-3-Sonnet &  &  &\underline{82.5} & 80.0 &  & \cellcolor{red!35}\underline{80.5} & 73.0 &  & 55.0 & 56.0 &  & \underline{40.3} & 35.3 &  & \underline{62.7} & 58.5 \\
Gemini-1.5-Flash &  &  &\underline{85.0} & 78.0 &  & \underline{78.5} & 69.5 &  & \underline{55.0} & 46.0 &  & \underline{40.0} & 31.7 &  & \cellcolor{red!5}\underline{62.8} & 54.5 \\
Gemini-1.5-Pro &  &  &\cellcolor{red!20}\underline{85.5} & \cellcolor{red!5}80.5 &  & \cellcolor{red!20}\underline{80.0} & 58.0 &  & \cellcolor{red!35}\underline{58.0} & 55.0 &  & \underline{40.3} & 30.0 &  & \cellcolor{red!35}\underline{63.7} & 52.8 \\
Claude-3-Haiku &  &  &74.5 & 79.0 &  & \underline{71.5} & 58.5 &  & \underline{55.0} & 50.0 &  & \underline{36.7} & 31.7 &  & \underline{57.1} & 52.5 \\
GPT-4o-Mini &  &  &\cellcolor{red!35}\underline{88.5} & 69.5 &  & \underline{77.0} & 69.5 &  & 53.0 & 56.0 &  & \underline{38.7} & 28.0 &  & \underline{62.5} & 52.2 \\
GPT-3.5-Turbo &  &  &\underline{71.0} & 60.5 &  & \underline{52.5} & 39.0 &  & \underline{41.0} & 28.0 &  & \underline{28.7} & 15.0 &  & \underline{46.8} & 34.0 \\

\midrule
\multicolumn{18}{c}{\emph{\textbf{Open-source LLMs}}} \\\noalign{\vskip 0.5ex}

DeepSeek-V2 & 236B & MoE &\cellcolor{red!35}\underline{87.0} & \cellcolor{red!20}82.0 &  & \cellcolor{red!5}\underline{75.5} & 69.5 &  & \cellcolor{red!35}\underline{61.0} & \cellcolor{red!35}56.0 &  & \cellcolor{red!35}\underline{43.0} & \cellcolor{red!35}39.7 &  & \cellcolor{red!35}\underline{64.4} & \cellcolor{red!35}59.8 \\
Mistral-Large & 123B &  &\cellcolor{red!5}\underline{85.0} & \cellcolor{red!35}83.5 &  & \cellcolor{red!20}76.5 & \cellcolor{red!35}81.0 &  & \cellcolor{red!5}\underline{56.0} & \cellcolor{red!20}55.0 &  & \underline{41.0} & \cellcolor{red!5}31.3 &  & \cellcolor{red!5}\underline{62.8} & \cellcolor{red!20}59.7 \\
DeepSeek-Coder-V2 & 236B & Code, MoE &\cellcolor{red!20}\underline{85.0} & \cellcolor{red!5}79.0 &  & \cellcolor{red!35}\underline{78.0} & 66.5 &  & \cellcolor{red!20}\underline{56.0} & \cellcolor{red!5}54.0 &  & \cellcolor{red!20}\underline{41.0} & \cellcolor{red!20}37.7 &  & \cellcolor{red!20}\underline{63.1} & \cellcolor{red!5}57.3 \\
Llama-3.1 & 70B &  &74.5 & 76.5 &  & 68.0 & \cellcolor{red!5}71.0 &  & \underline{53.0} & 50.0 &  & \underline{34.7} & 29.3 &  & \underline{55.3} & 54.1 \\
Qwen2 & 72B &  &26.5 & 74.0 &  & 24.5 & \cellcolor{red!20}72.5 &  & 8.0 & 45.0 &  & 7.0 & 27.0 &  & 16.4 & 52.4 \\
Llama-3 & 70B &  &\underline{84.5} & 73.5 &  & \underline{64.0} & 63.5 &  & \underline{52.0} & 42.0 &  & \cellcolor{red!5}\underline{41.0} & 28.3 &  & \underline{59.0} & 50.1 \\
Mixtral-8x22B & 141B & MoE &30.0 & 74.0 &  & 21.5 & 57.0 &  & 25.0 & 47.0 &  & 14.7 & 24.0 &  & 21.5 & 47.6 \\
Gemma-2 & 9B &  &\underline{79.0} & 66.5 &  & \underline{65.0} & 54.5 &  & \underline{50.0} & 39.0 &  & \underline{24.3} & 17.7 &  & \underline{51.4} & 41.8 \\
DeepSeek-Coder-V2-Lite & 16B & Code &66.0 & 67.5 &  & 51.0 & 53.5 &  & 27.0 & 30.0 &  & \underline{22.0} & 20.3 &  & 40.9 & 41.6 \\
WizardLM-2 & 141B & MoE &\underline{62.5} & 60.5 &  & \underline{56.5} & 55.5 &  & 25.0 & 34.0 &  & 17.7 & 18.0 &  & 39.5 & 40.0 \\
C4AI Command R+ & 104B &  &35.5 & 65.5 &  & 39.0 & 51.0 &  & 19.0 & 31.0 &  & 8.7 & 18.3 &  & 24.3 & 39.9 \\
Yi-1.5 & 9B &  &18.0 & 68.5 &  & 24.5 & 56.0 &  & 2.0 & 14.0 &  & 4.0 & 14.0 &  & 12.4 & 38.1 \\
Yi-1.5 & 34B &  &0.5 & 64.5 &  & 1.0 & 53.0 &  & 0.0 & 14.0 &  & 0.0 & 15.3 &  & 0.4 & 36.9 \\
Mistral-Nemo & 12B &  &52.5 & 59.5 &  & 37.5 & 44.0 &  & 28.0 & 37.0 &  & 15.3 & 16.7 &  & 31.7 & 36.8 \\
Llama-3.1 & 8B &  &\underline{62.0} & 60.0 &  & \underline{44.0} & 42.5 &  & 32.0 & 33.0 &  & \underline{19.0} & 14.3 &  & \underline{37.6} & 35.1 \\
DBRX & 132B & MoE &41.0 & 57.0 &  & 29.5 & 43.0 &  & \underline{32.0} & 30.0 &  & 12.0 & 16.3 &  & 26.1 & 34.9 \\
Codestral & 22B & Code &39.0 & 51.5 &  & 38.5 & 41.5 &  & 18.0 & 23.0 &  & \underline{17.3} & 13.0 &  & 28.1 & 31.0 \\
Llama-3 & 8B &  &49.5 & 56.5 &  & 21.5 & 31.0 &  & 24.0 & 29.0 &  & 10.0 & 12.3 &  & 24.5 & 30.1 \\
Qwen2 & 7B &  &13.0 & 56.0 &  & 9.5 & 33.0 &  & 4.0 & 31.0 &  & 2.3 & 10.0 &  & 7.0 & 29.9 \\
Mathstral & 7B & Math &43.5 & 55.0 &  & 32.5 & 35.0 &  & 10.0 & 23.0 &  & 11.3 & 11.7 &  & 24.5 & 29.8 \\
GLM-4 & 9B &  &\underline{69.5} & 44.0 &  & \underline{53.5} & 34.0 &  & \underline{33.0} & 20.0 &  & \underline{17.7} & 8.7 &  & \underline{41.5} & 25.3 \\
Aya-23 & 35B &  &1.5 & 44.0 &  & 1.0 & 25.5 &  & 0.0 & 20.0 &  & 0.0 & 11.7 &  & 0.6 & 24.3 \\
DeepSeek-V2-Lite & 16B & MoE &7.0 & 45.5 &  & 3.5 & 18.0 &  & 1.0 & 17.0 &  & 1.0 & 10.3 &  & 3.1 & 21.9 \\
Mixtral-8x7B-v0.1 & 46B & MoE &0.5 & 39.0 &  & 2.0 & 17.0 &  & 0.0 & 25.0 &  & 0.0 & 12.7 &  & 0.6 & 21.9 \\
DeepSeek-Math & 7B & Math &2.0 & 46.0 &  & 1.0 & 27.0 &  & 1.0 & 4.0 &  & 0.3 & 8.0 &  & 1.0 & 21.8 \\
Llama-2 & 70B &  &32.5 & 43.5 &  & 16.5 & 25.0 &  & 1.0 & 8.0 &  & 2.0 & 7.0 &  & 13.1 & 20.8 \\
WizardLM-2 & 7B &  &\underline{47.0} & 42.0 &  & \underline{30.5} & 28.5 &  & 5.0 & 6.0 &  & \underline{7.3} & 5.7 &  & \underline{22.7} & 20.5 \\
Mistral-v0.3 & 7B &  &\underline{49.5} & 40.0 &  & \underline{40.5} & 28.0 &  & \underline{25.0} & 9.0 &  & \underline{11.3} & 5.7 &  & \underline{29.9} & 20.3 \\
WizardMath & 7B & Math &22.5 & 32.0 &  & 12.0 & 22.5 &  & 6.0 & 7.0 &  & \underline{3.7} & 3.3 &  & 10.8 & 15.7 \\
InternLM2-Math-Plus & 7B & Math &\underline{28.5} & 27.5 &  & \underline{15.0} & 14.0 &  & 7.0 & 9.0 &  & \underline{4.7} & 4.0 &  & \underline{13.5} & 13.0 \\
StarCoder2 & 15B & Code &\underline{47.5} & 21.0 &  & \underline{34.0} & 15.5 &  & \underline{11.0} & 6.0 &  & \underline{8.3} & 4.3 &  & \underline{24.9} & 11.5 \\
InternLM2 & 7B &  &18.0 & 20.0 &  & 4.5 & 11.0 &  & 9.0 & 10.0 &  & \underline{2.7} & 2.3 &  & 7.8 & 9.9 \\
Gemma-1 & 7B &  &1.0 & 20.0 &  & 0.0 & 7.5 &  & 0.0 & 7.0 &  & 0.0 & 3.3 &  & 0.2 & 9.0 \\
Llama-2 & 7B &  &4.0 & 17.0 &  & 4.0 & 11.5 &  & 0.0 & 2.0 &  & 1.3 & 2.7 &  & 2.5 & 8.4 \\
DeepSeek-Coder-V1 & 33B & Code &\underline{19.0} & 18.5 &  & 8.5 & 8.5 &  & 2.0 & 2.0 &  & \underline{3.7} & 1.7 &  & \underline{8.5} & 7.6 \\
WizardCoder & 33B & Code &\underline{32.5} & 16.0 &  & \underline{17.5} & 8.0 &  & \underline{5.0} & 2.0 &  & \underline{5.0} & 1.0 &  & \underline{15.0} & 6.6 \\
Aya-23 & 8B &  &1.0 & 13.0 &  & 0.0 & 9.0 &  & 0.0 & 2.0 &  & 0.3 & 2.3 &  & 0.4 & 6.6 \\
Gemma-1 & 2B &  &4.0 & 8.0 &  & 1.5 & 7.5 &  & 0.0 & 2.0 &  & 0.0 & 0.0 &  & 1.4 & 4.1 \\

\bottomrule
\end{tabular}
}
\caption{LLM performance on the \emph{testmini} set of \ours.
We utilize the average accuracy achieved through CoT prompting as the metric for ranking model performance.
For \ourcl, we use the OpenAI Embedding 3 Large retriever to retrieve top-$10$ evidence as input document.
Numbers underlined indicate that models using PoT prompting outperform those using CoT prompting.
}
\label{tab:testmini}
\end{table*}

%% file: tables/retrieval.tex
\begin{table}[!t]
    \centering
    \addtolength{\tabcolsep}{-0.2em}
    \renewcommand{\arraystretch}{1.1}
    \small
    \begin{tabular}{clrrr}
        \toprule
        top-$n$ & Retriever & R@$n$ & Llama-3 & GPT-4o \\
        \midrule
        \multirow{4}{*}{3}
        & Contriever & 22.3 & 13.7 & 16.0\\
        & BM25 & 29.5 & 13.7 & 15.7\\
        & Embedding-3-Small & 44.7 & 19.0 & 24.0\\
        & Embedding-3-Large & 48.2 & 22.0 & 27.0\\
        \noalign{\vskip 0.5ex}\hdashline\noalign{\vskip 0.5ex}

        \multirow{4}{*}{5} 
        & Contriever & 32.0 & 15.3 & 22.0\\
        & BM25 & 38.0 & 15.3 & 20.7 \\
        & Embedding-3-Small & 57.1 & 21.0 & 29.0\\
        & Embedding-3-Large & 62.0 & 24.0 & 32.7\\
        \noalign{\vskip 0.5ex}\hdashline\noalign{\vskip 0.5ex}
        
        \multirow{4}{*}{10} 
        & Contriever & 45.3 & 18.3 & 25.7\\
        & BM25 & 47.9 & 20.3 & 23.0\\
        & Embedding-3-Small & 71.2 & 25.3 & 31.7\\
        & Embedding-3-Large & 75.8 & 26.3 & 36.7 \\
        \midrule

        -- & Oracle &  & 35.3 & 42.0 \\
        \bottomrule
    \end{tabular}
    
    \caption{Results of the Llama-3-70B and GPT-4o with CoT prompting approaches under various retrieval settings on the \ourcl testmini set. A correlation is observed between LLM performance and the question-relevance of the retrieved evidence.}
    \label{tab:retriever}
\end{table}

%% file: tables/long_context.tex
\begin{table}[!t]
\centering
\footnotesize
\resizebox{0.85\linewidth}{!}{
\renewcommand{\arraystretch}{1.1}
\begin{tabular}{lrr}

    \toprule
    Model &            RAG &   Long Context \\
    \midrule
    GPT-4o &          36.7& \textbf{40.3} \\
    Gemini-1.5-Pro &           30.0 & \textbf{37.3} \\
    Claude-3-Sonnet   & \textbf{35.3} &          34.7 \\
    Gemini-1.5-Flash &          31.7 & \textbf{34.3} \\
    Claude-3-Haiku  & \textbf{31.7} &           31.0 \\
    \noalign{\vskip 0.5ex}\hdashline\noalign{\vskip 0.5ex}
    DeepSeek-V2 & \textbf{39.7} &          38.7 \\
    DeepSeek-Coder-V2 & \textbf{37.7} &           36.0 \\
    Llama-3.1-70B & \textbf{29.3} & 26.3 \\
    Llama-3.1-8B & \textbf{14.3} & 9.0\\
    Mistral-Nemo & \textbf{16.7} & 4.7 \\
    Phi-3-Medium    &          12.7 &  \textbf{13.0} \\
    GLM-4-9B &           8.7 &  \textbf{9.7} \\
    \bottomrule
\end{tabular}
}
\caption{Results of the CoT prompting approach under various retrieval settings on \ourcl testmini set.}

\label{tab:long-context-results}
\end{table}

%% file: tables/case_study.tex
\begin{table}[!t]
\small
\centering
\begin{tabular}{p{2.5cm}p{4.3cm}}
\toprule
\textbf{Error Type} & \textbf{Explanation} \\
\midrule

\noalign{\vspace{0.5em}}

Inaccurate Evidence Retrieval \newline (39 / 100) & The challenge lies in finding accurate evidence, especially in situations where the values needed for intermediate reasoning steps are not explicitly stated. This makes it difficult for the retriever to identify the correct evidence. \\

\noalign{\vspace{0.5em}}\hdashline\noalign{\vspace{0.5em}}

Calculation Error \newline (28 / 100)  & The reasoning process is accurate, but there are errors in the intermediate or final computations. \\

\noalign{\vspace{0.5em}}\hdashline\noalign{\vspace{0.5em}}

Table Misunderstanding \newline (16 / 100) & The model faces challenges in comprehending and parsing cell values, particularly in complex tables.\\

\noalign{\vspace{0.5em}}\hdashline\noalign{\vspace{0.5em}}

Exceeding Context Length (8 / 100) & The input document exceeds the context length limit.\\

\noalign{\vspace{0.5em}}\hdashline\noalign{\vspace{0.5em}}

Others &   \\
\bottomrule
\end{tabular}
\caption{Error types and explanations of GPT-3.5-turbo failure cases on the \ourcl testmini set.}
\label{tab:casestudy}
\end{table}

%% file: main/6-conclusion.tex
\section{Conclusion}
This paper introduces \ours, a comprehensive benchmark designed to evaluate the capabilities of LLMs in numerical reasoning over long and \spe documents. 
Our experiments show that even the best-performing current models still fall short of human expert performance on problems requiring complex reasoning over extended contexts. 
This highlights the need for future research to improve LLMs' proficiency in complex numerical reasoning tasks within expert domains.

%% file: main/limitations.tex
\section*{Limitations}
There are some limitations in our study that we believe can be addressed in future work. First, our approach to extracting the final answer from the model's output is not yet flawless. In certain instances, this method fails to accurately identify the answer, causing the reported accuracy to be an approximate lower limit.
Additionally, we suggest that future research could investigate training large language models (LLMs) on finance-specific data to improve their performance on the \ours benchmark~\cite{Wu2023BloombergGPTAL, luukkonen-etal-2023-fingpt, Xie2023PIXIUAL}.

%% file: main/acknowledgement.tex
\section*{Acknowledgement}
We are grateful for the compute support provided by Microsoft Research’s Accelerate Foundation Models Research (AFMR) program.
We extend our gratitude to the anonymous reviewers and area chairs for their valuable discussions and feedback.

%% file: appendix/main.tex
\newpage
\section{Appendix}
\input{tables/annotator_agreement}
\input{figures/PoT_prompt_example}
\input{tables/annotator_info}
\input{tables/model_note}
\input{tables/test_results}

%% file: tables/annotator_agreement.tex
\begin{table}[h]
\centering
\small
\addtolength{\tabcolsep}{-0.1em}
\begin{tabular}{lc}
\toprule
\textbf{Annotation Quality}    & \textbf{\%S $\geq$ 4}  \\
\midrule
Question Fluency  & 97.4\\
Question Correctness &  96.0\\
\noalign{\vskip 0.5ex}\hdashline\noalign{\vskip 0.5ex}
Evidence Relevance  &  88.5\\
Evidence Completeness & 91.3\\
\noalign{\vskip 0.5ex}\hdashline\noalign{\vskip 0.5ex}
Final Answer Correctness & 97.9\\
Python Solution Correctness & 97.6\\
Variable Value Correctness & 98.5\\
Python Solution Conciseness & 89.1\\
Variable Name Meaningfulness & 95.4\\
\bottomrule
\end{tabular}
\caption{Human evaluation was conducted on 200 samples from \ours, with three internal reviewers asked to rate each sample on a scale from 1 to 5. We present the percentage of samples that received an average score of 4 or higher, as an indicator of the annotation quality of \ours.}
\label{tab:annotation_aggrement}
\end{table}

%% file: figures/PoT_prompt_example.tex
\begin{figure}[h]
    \centering
    \includegraphics[width = 0.8 \linewidth]{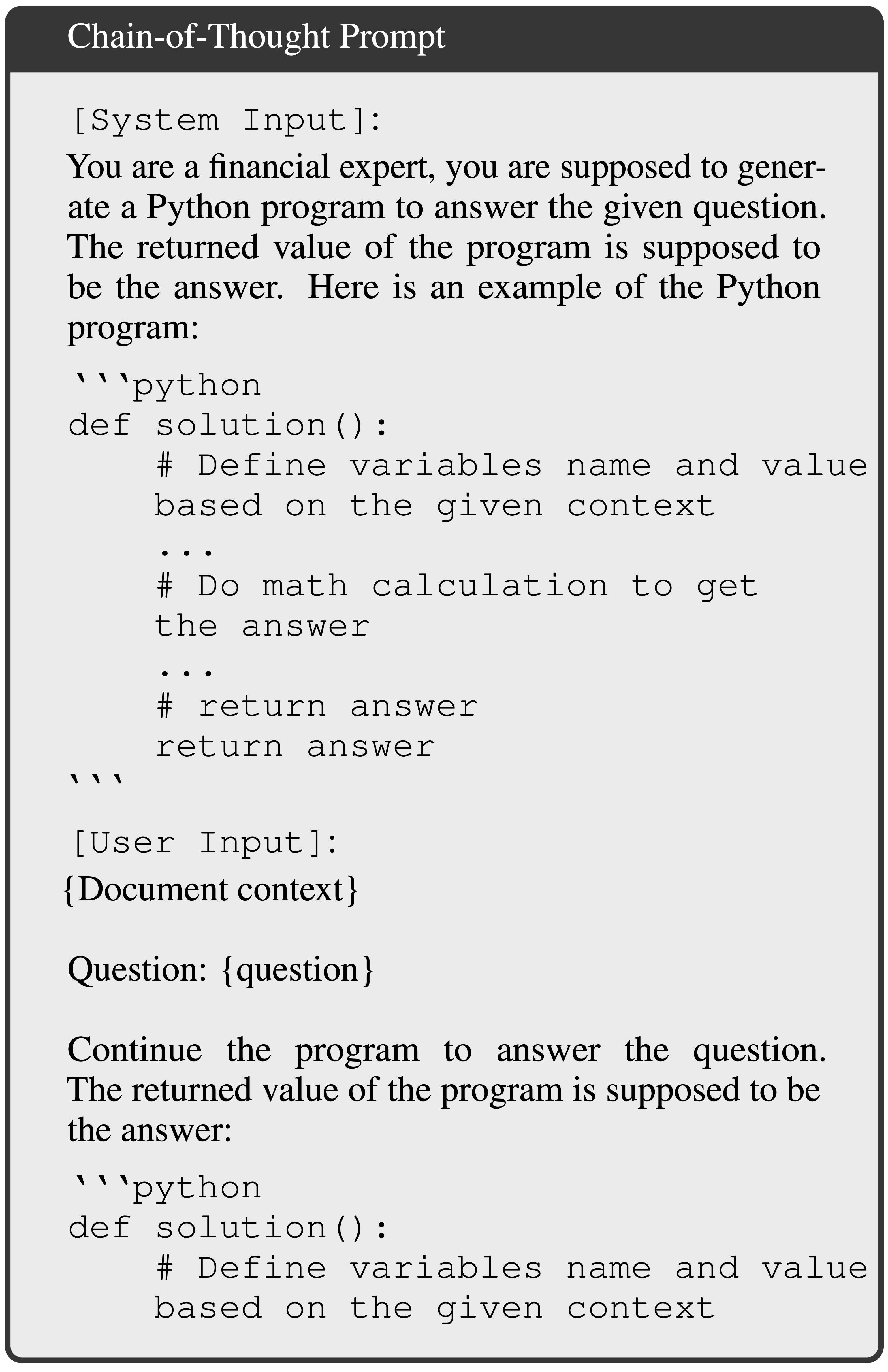}

\caption{Example of \emph{zero}-shot PoT prompt used.}
\label{fig:pot_example}
\end{figure}

%% file: tables/annotator_info.tex
\begin{table*}[!t]
\centering
\small
\resizebox{\textwidth}{!}{%
\begin{tabular}{clllc}
\toprule
\textbf{Annotator ID} & \textbf{Finance Industry Experience} & \textbf{Annotation Sets}\\
\midrule
1 & 1 working and 1 internship at US &  New subset, Annotation validation \\

2 & >= 2 internship at US & New subset, Annotation validation \\

3 & 1 working at Singapore and 2 internship at US &  New subset \\

4 & 2 working and >= 1 internship at US  &  New subset \\

5 & 1 internship at US, 2 internship at China & Re-annotation on three subsets, Annotation validation \\

6 & Graduate student majored in computer science & Re-annotation on three subsets, Annotation validation \\

7 & Graduate student majored in statistics & Re-annotation on three subsets \\
\bottomrule
\end{tabular}
}
\caption{Details of annotators involved in dataset construction.}
\label{tab:annotator_profiles}
\end{table*}

%% file: tables/model_note.tex
\begin{table*}[!t]
\centering
\resizebox{\textwidth}{!}{%
\renewcommand{\arraystretch}{1.1}
\begin{tabular}{llllp{10cm}}
\toprule
Organization & Model & Size & Notes & Source \\
\midrule
\multirow{3}{*}{OpenAI} & GPT-4-Turbo & -- & & \texttt{gpt-4o-2024-05-13} \\
 & GPT-4o & -- & & \texttt{gpt-4-turbo-2024-04-09} \\
 & GPT-3.5-Turbo & -- & & \texttt{gpt-3.5-turbo-0125} \\
\noalign{\vskip 0.5ex}\hdashline\noalign{\vskip 0.5ex}

\multirow{4}{*}{Anthropic} & Claude-3.5-Sonnet & -- & & \texttt{claude-3-5-sonnet-20240620} \\
 & Claude-3-Opus & -- & & \texttt{claude-3-opus-20240229} \\
 & Claude-3-Sonnet & -- & & \texttt{claude-3-sonnet-20240229} \\
 & Claude-3-Haiku & -- & & \texttt{claude-3-haiku-20240307} \\
\noalign{\vskip 0.5ex}\hdashline\noalign{\vskip 0.5ex}

\multirow{2}{*}{Google} & Gemini-1.5-Pro & -- & & \texttt{gemini-1.5-pro} \\
 & Gemini-1.5-Flash & -- & & \texttt{gemini-1.5-flash} \\

\midrule \midrule
\multirow{1}{*}{Alibaba} & Qwen2 & 7 \& 72B & & \texttt{Qwen/Qwen2-*B-Instruct} \\

\noalign{\vskip 0.5ex}\hdashline\noalign{\vskip 0.5ex}
\multirow{3}{*}{Meta} & Llama-2 & 7 \& 70B & & \texttt{meta-llama/Llama-2-*b-chat-hf} \\
 & Llama-3 & 8 \& 70B & & \texttt{meta-llama/Meta-Llama-3-*B-Instruct} \\
 & Llama-3.1 & 8 \& 70B \& 405B & & \texttt{meta-llama/Meta-Llama-3.1-*B-Instruct} \\

\noalign{\vskip 0.5ex}\hdashline\noalign{\vskip 0.5ex}

\multirow{2}{*}{Google} & Gemma-1 & 2 \& 7B & & \texttt{google/gemma-b-it} \\
 & Gemma-2 & 9B & & \texttt{google/gemma-2-9b-it} \\
 \noalign{\vskip 0.5ex}\hdashline\noalign{\vskip 0.5ex}
 
\multirow{6}{*}{Mistral AI} & Mistral-v0.3 & 7B & & \texttt{mistralai/Mistral-7B-Instruct-v0.3} \\
 & Mistral-Nemo & 12B & & \texttt{mistralai/Mistral-Nemo-Instruct-2407} \\
 & Mistral-Large & 123B & & \texttt{mistralai/Mistral-Large-Instruct-2407} \\
 & Mathstral & 7B & Math-Specific & \texttt{mistralai/Mathstral-7B-v0.1} \\
 & Mixtral & 46 \& 141B & MoE & \texttt{mistralai/Mixtral--Instruct-v0.1} \\
 & Codestral & 22B & Code-Specific & \texttt{mistralai/Codestral-22B-v0.1} \\

\noalign{\vskip 0.5ex}\hdashline\noalign{\vskip 0.5ex}
\multirow{4}{*}{DeepSeek} & DeepSeek-Math & 7B & Math-Specific & \texttt{deepseek-ai/deepseek-math-7b-instruct} \\
 & DeepSeek-Coder-V1 & 33B & Code-Specific & \texttt{deepseek-ai/deepseek-coder-33b-instruct} \\
 & DeepSeek-V2 & 16 \& 236B & MoE & \texttt{deepseek-ai/DeepSeek-V2-*-Chat}\\
 & DeepSeek-Coder-V2 & 16 \& 236B & Code-Specific, MoE & \texttt{deepseek-ai/DeepSeek-Coder-V2-*-Instruct} \\

\noalign{\vskip 0.5ex}\hdashline\noalign{\vskip 0.5ex}
\multirow{1}{*}{01 AI} & Yi-1.5 & 9 \& 34B & & \texttt{01-ai/Yi-1.5-34B-Chat} \\

\noalign{\vskip 0.5ex}\hdashline\noalign{\vskip 0.5ex}
\multirow{2}{*}{Microsoft} & Phi-3-Medium & 14B & & \texttt{microsoft/Phi-3-medium-4k-instruct} \\
 & Phi-3-Mini & 3B & & \texttt{microsoft/Phi-3-mini-4k-instruct} \\

\noalign{\vskip 0.5ex}\hdashline\noalign{\vskip 0.5ex}
\multirow{1}{*}{THUDM} & GLM-4 & 9B & & \texttt{THUDM/glm-4-9b-chat} \\

\noalign{\vskip 0.5ex}\hdashline\noalign{\vskip 0.5ex}
\multirow{1}{*}{Databricks} & DBRX & 132B & MoE & \texttt{databricks/dbrx-instruct} \\

\noalign{\vskip 0.5ex}\hdashline\noalign{\vskip 0.5ex}
\multirow{1}{*}{Cohere} & C4AI Command R+ & 104B & & \texttt{CohereForAI/c4ai-command-r-plus} \\
 & Aya-23 & 8 \& 35B & & \texttt{CohereForAI/aya-23-*B} \\

\noalign{\vskip 0.5ex}\hdashline\noalign{\vskip 0.5ex}
\multirow{2}{*}{InternLM} & InternLM2 & 7B & & \texttt{internlm/internlm2-chat-7b} \\
 & InternLM2-Math-Plus & 7B & Math-Specific & \texttt{internlm/internlm2-math-plus-7b} \\

\noalign{\vskip 0.5ex}\hdashline\noalign{\vskip 0.5ex}
\multirow{4}{*}{WizardLM Team} & WizardLM-2 & 7B & & \texttt{lucyknada/microsoft\_WizardLM-2-7B} \\
 & WizardMath & 7B & Math-Specific & \texttt{WizardLMTeam/WizardMath-7B-V1.1} \\
 & WizardCoder & 33B & Code-Specific & \texttt{WizardLMTeam/WizardCoder-33B-V1.1} \\
 & WizardLM-2 (MoE) & 141B & MoE & \texttt{alpindale/WizardLM-2-8x22B} \\

\noalign{\vskip 0.5ex}\hdashline\noalign{\vskip 0.5ex}
\multirow{1}{*}{BigCode} & StarCoder2 & 15B & Code-Specific & \texttt{bigcode/starcoder2-15b-instruct-v0.1} \\

\bottomrule
\end{tabular}
}
\caption{Details of the LLMs evaluated in this study. }
\label{tab:model-detail}
\end{table*}

%% file: tables/test_results.tex
\begin{table*}[!t]
\centering
\resizebox{\textwidth}{!}{%
\addtolength{\tabcolsep}{0.1em}
\begin{tabular}{lrlrrrrrrrrrrrrrrr}
\toprule
\multirow{2}{*}{\textbf{Model}} & \multirow{2}{*}{\textbf{Size}} & \multirow{2}{*}{\textbf{Notes}} & \multicolumn{2}{c}{\textbf{\ourss}} && \multicolumn{2}{c}{\textbf{\ourcs}} && \multicolumn{2}{c}{\textbf{\oursl}} && \multicolumn{2}{c}{\textbf{\ourcl}} && \multicolumn{2}{c}{\textbf{Avg. Acc}} \\
\cmidrule(lr){4-5} \cmidrule(lr){7-8} \cmidrule(lr){10-11} \cmidrule(lr){13-14} \cmidrule(lr){16-17}
& & & PoT & CoT && PoT & CoT && PoT & CoT && PoT & CoT &&  PoT & CoT\\
\midrule

Human Expert & & & \multicolumn{3}{c}{91.0} & \multicolumn{3}{c}{87.0} & \multicolumn{3}{c}{84.0} & \multicolumn{3}{c}{76.0} &\\
\midrule

\multicolumn{18}{c}{\emph{\textbf{Proprietary LLMs}}} \\\noalign{\vskip 0.5ex}

GPT-4-Turbo &  &  &\cellcolor{red!35}\underline{88.9} & \cellcolor{red!20}86.2 &  & \underline{78.6} & \cellcolor{red!35}77.8 &  & 31.5 & \cellcolor{red!5}61.2 &  & 20.2 & \cellcolor{red!35}35.6 &  & 53.4 & \cellcolor{red!35}62.0 \\
GPT-4o &  &  &\underline{87.0} & \cellcolor{red!35}86.4 &  & 69.6 & \cellcolor{red!20}75.9 &  & \cellcolor{red!5}62.2 & \cellcolor{red!35}66.2 &  & \cellcolor{red!20}\underline{41.5} & \cellcolor{red!20}31.7 &  & \underline{62.5} & \cellcolor{red!20}60.7 \\
Claude-3.5-Sonnet &  &  &\underline{87.2} & 81.8 &  & \underline{71.0} & 70.0 &  & \cellcolor{red!35}\underline{65.0} & \cellcolor{red!20}61.5 &  & \cellcolor{red!5}\underline{40.8} & \cellcolor{red!5}31.5 &  & \cellcolor{red!5}\underline{63.0} & \cellcolor{red!5}57.4 \\
GPT-4o-Mini &  &  &\cellcolor{red!20}\underline{88.8} & 76.0 &  & \underline{77.2} & \cellcolor{red!5}72.4 &  & \underline{56.8} & 50.2 &  & \underline{36.9} & 29.2 &  & \underline{62.4} & 54.3 \\
Gemini-1.5-Pro &  &  &\underline{87.9} & 82.5 &  & \cellcolor{red!35}\underline{80.4} & 58.4 &  & \cellcolor{red!20}\underline{63.0} & 58.0 &  & \cellcolor{red!35}\underline{41.8} & 31.2 &  & \cellcolor{red!35}\underline{65.6} & 54.2 \\
Claude-3-Haiku &  &  &77.8 & 82.2 &  & \underline{72.4} & 61.5 &  & \underline{55.0} & 53.2 &  & \underline{34.6} & 30.8 &  & \underline{57.4} & 54.2 \\
Gemini-1.5-Flash &  &  &\cellcolor{red!5}\underline{87.9} & 80.5 &  & \cellcolor{red!5}\underline{79.9} & 67.0 &  & \underline{60.5} & 53.8 &  & \underline{38.9} & 28.1 &  & \cellcolor{red!20}\underline{64.1} & 54.1 \\
GPT-3.5-Turbo &  &  &\underline{75.6} & 64.0 &  & \underline{47.6} & 34.2 &  & \underline{46.2} & 38.8 &  & \underline{23.2} & 13.2 &  & \underline{45.3} & 34.3 \\

\midrule
\multicolumn{18}{c}{\emph{\textbf{Open-source LLMs}}} \\\noalign{\vskip 0.5ex}

Mistral-Large & 123B &  &\cellcolor{red!5}\underline{87.9} & \cellcolor{red!20}85.0 &  & \cellcolor{red!5}74.4 & \cellcolor{red!35}79.8 &  & \cellcolor{red!35}58.8 & \cellcolor{red!35}58.8 &  & \cellcolor{red!5}\underline{37.2} & \cellcolor{red!5}30.6 &  & \cellcolor{red!5}\underline{61.8} & \cellcolor{red!35}60.0 \\
DeepSeek-V2 & 236B & MoE &\cellcolor{red!35}\underline{88.9} & \cellcolor{red!35}86.1 &  & \cellcolor{red!35}\underline{76.9} & 66.6 &  & \cellcolor{red!20}\underline{58.2} & \cellcolor{red!20}57.2 &  & \cellcolor{red!35}\underline{40.2} & \cellcolor{red!35}32.0 &  & \cellcolor{red!35}\underline{63.8} & \cellcolor{red!20}57.3 \\
DeepSeek-Coder-V2 & 236B & Code, MoE &\cellcolor{red!20}\underline{88.2} & \cellcolor{red!5}80.9 &  & \cellcolor{red!20}\underline{74.9} & 67.2 &  & \cellcolor{red!5}\underline{56.8} & \cellcolor{red!5}54.8 &  & \cellcolor{red!20}\underline{37.5} & \cellcolor{red!20}31.5 &  & \cellcolor{red!20}\underline{61.9} & \cellcolor{red!5}55.7 \\
Llama-3.1 & 70B &  &\underline{81.0} & 80.1 &  & 65.8 & \cellcolor{red!20}70.8 &  & 49.2 & 53.8 &  & \underline{32.1} & 26.6 &  & \underline{54.9} & 54.4 \\
Qwen2 & 72B &  &27.8 & 77.5 &  & 25.0 & \cellcolor{red!5}70.1 &  & 16.8 & 49.0 &  & 5.7 & 25.1 &  & 17.4 & 52.4 \\
Llama-3 & 70B &  &\underline{86.2} & 80.9 &  & \underline{64.8} & 62.0 &  & \underline{51.5} & 45.0 &  & \underline{35.0} & 26.3 &  & \underline{57.3} & 51.2 \\
Mixtral-8x22B & 141B & MoE &27.4 & 74.9 &  & 23.1 & 59.4 &  & 21.5 & 46.8 &  & 14.8 & 22.3 &  & 20.9 & 47.8 \\
DeepSeek-Coder-V2-Lite & 16B & Code &68.8 & 71.4 &  & \underline{52.2} & 44.9 &  & 28.8 & 35.0 &  & 18.0 & 19.8 &  & 40.6 & 40.9 \\
Gemma-2 & 9B &  &\underline{82.0} & 70.1 &  & \underline{62.4} & 49.2 &  & \underline{45.5} & 33.2 &  & \underline{24.4} & 17.7 &  & \underline{50.9} & 40.6 \\
Yi-1.5 & 34B &  &1.0 & 71.2 &  & 0.9 & 58.2 &  & 0.2 & 17.8 &  & 0.2 & 13.7 &  & 0.6 & 39.7 \\
C4AI Command R+ & 104B &  &37.6 & 67.5 &  & 36.5 & 50.6 &  & 19.2 & 35.0 &  & 7.6 & 15.1 &  & 23.8 & 39.6 \\
WizardLM-2 & 141B & MoE &\underline{62.1} & 59.4 &  & \underline{49.5} & 47.6 &  & 31.2 & 36.0 &  & \underline{19.0} & 16.1 &  & \underline{38.9} & 37.3 \\
DBRX & 132B & MoE &46.8 & 64.0 &  & 33.1 & 41.5 &  & 26.5 & 29.8 &  & 10.7 & 18.4 &  & 27.3 & 37.0 \\
Mistral-Nemo & 12B &  &51.1 & 66.0 &  & 32.1 & 40.5 &  & 29.0 & 37.0 &  & \underline{15.7} & 15.0 &  & 30.3 & 36.9 \\
Yi-1.5 & 9B &  &23.2 & 69.5 &  & 17.4 & 45.0 &  & 1.2 & 15.0 &  & 2.2 & 11.9 &  & 11.1 & 35.0 \\
Llama-3.1 & 8B &  &\underline{66.5} & 63.0 &  & \underline{41.9} & 34.1 &  & 35.5 & 35.5 &  & \underline{14.7} & 13.8 &  & \underline{37.0} & 33.9 \\
Codestral & 22B & Code &43.1 & 58.0 &  & 36.8 & 37.2 &  & 25.0 & 29.5 &  & \underline{14.2} & 14.0 &  & 28.4 & 32.8 \\
Llama-3 & 8B &  &51.6 & 57.8 &  & 22.8 & 31.2 &  & 22.5 & 25.0 &  & 9.3 & 11.5 &  & 24.9 & 29.7 \\
Mathstral & 7B & Math &45.8 & 54.4 &  & 31.1 & 34.0 &  & 11.0 & 24.0 &  & 9.1 & 11.7 &  & 24.0 & 29.5 \\
Qwen2 & 7B &  &15.4 & 52.6 &  & 6.6 & 33.2 &  & 4.2 & 29.0 &  & 2.6 & 11.4 &  & 7.0 & 29.4 \\
GLM-4 & 9B &  &\underline{68.0} & 48.8 &  & \underline{46.5} & 32.1 &  & \underline{32.5} & 22.8 &  & \underline{18.2} & 11.1 &  & \underline{39.5} & 27.2 \\
Aya-23 & 35B &  &0.9 & 46.9 &  & 0.5 & 26.5 &  & 0.0 & 19.5 &  & 0.5 & 10.1 &  & 0.5 & 24.6 \\
Mixtral-8x7B-v0.1 & 46B & MoE &1.1 & 42.5 &  & 0.4 & 19.6 &  & 0.2 & 24.2 &  & 0.2 & 13.0 &  & 0.5 & 23.4 \\
DeepSeek-Math & 7B & Math &1.4 & 47.1 &  & 0.4 & 28.0 &  & 1.2 & 12.5 &  & 0.5 & 7.9 &  & 0.8 & 23.3 \\
Mistral-v0.3 & 7B &  &\underline{48.6} & 40.8 &  & \underline{28.5} & 24.8 &  & \underline{19.5} & 18.0 &  & \underline{12.8} & 7.6 &  & \underline{26.5} & 21.5 \\
DeepSeek-V2-Lite & 16B & MoE &7.6 & 49.1 &  & 4.6 & 16.9 &  & 2.5 & 15.2 &  & 1.0 & 8.3 &  & 3.7 & 21.5 \\
Llama-2 & 70B &  &27.1 & 45.1 &  & 16.8 & 26.0 &  & 2.0 & 8.0 &  & 1.1 & 6.9 &  & 11.6 & 21.4 \\
WizardLM-2 & 7B &  &\underline{47.0} & 42.6 &  & 31.2 & 31.2 &  & \underline{9.2} & 8.5 &  & \underline{7.2} & 4.8 &  & \underline{23.4} & 21.3 \\
WizardMath & 7B & Math &22.1 & 34.2 &  & 14.4 & 24.8 &  & 5.8 & 6.2 &  & 3.4 & 4.6 &  & 11.1 & 17.2 \\
InternLM2-Math-Plus & 7B & Math &30.0 & 30.2 &  & 15.0 & 16.2 &  & 10.8 & 11.2 &  & \underline{4.2} & 3.3 &  & 14.2 & 14.3 \\
StarCoder2 & 15B & Code &\underline{51.0} & 29.8 &  & \underline{32.0} & 16.8 &  & \underline{9.5} & 5.2 &  & \underline{8.4} & 3.0 &  & \underline{25.1} & 13.4 \\
InternLM2 & 7B &  &17.4 & 24.5 &  & 9.4 & 11.8 &  & \underline{9.0} & 6.8 &  & 2.9 & 4.3 &  & 8.9 & 11.5 \\
Gemma-1 & 7B &  &0.5 & 23.2 &  & 0.2 & 6.0 &  & 0.0 & 7.8 &  & 0.2 & 3.2 &  & 0.3 & 9.5 \\
Llama-2 & 7B &  &5.6 & 20.4 &  & 2.4 & 11.8 &  & 1.0 & 3.2 &  & 0.4 & 2.2 &  & 2.3 & 9.3 \\
WizardCoder & 33B & Code &\underline{38.4} & 19.9 &  & \underline{17.5} & 8.4 &  & \underline{7.2} & 3.2 &  & \underline{6.1} & 1.8 &  & \underline{17.2} & 8.2 \\
Aya-23 & 8B &  &0.5 & 14.2 &  & 0.1 & 8.5 &  & 0.0 & 2.8 &  & 0.0 & 2.2 &  & 0.2 & 6.8 \\
DeepSeek-Coder-V1 & 33B & Code &\underline{19.4} & 16.2 &  & \underline{8.5} & 6.6 &  & \underline{3.2} & 2.2 &  & \underline{3.6} & 1.2 &  & \underline{8.7} & 6.4 \\
Gemma-1 & 2B &  &5.9 & 9.1 &  & 2.5 & 6.0 &  & 0.8 & 2.8 &  & 0.2 & 1.0 &  & 2.3 & 4.5 \\

\bottomrule
\end{tabular}
}
\caption{Results of Chain-of-Thought and Program-of-Thought prompting on the \emph{test} set of \ours.
We use average Accuracy using CoT prompting as the ranking indicator of model performance. 
For \ourcl, we use the OpenAI Embedding 3 Large retriever to retrieve top-$10$ evidence as input document.
\underline{Numbers} underscored indicate that models with PoT prompting achieves better results than with CoT prompting.
}
\label{tab:test}
\end{table*}